%% file: IEEE-conference-template-062824.tex
\documentclass[conference]{IEEEtran}
\IEEEoverridecommandlockouts

\usepackage{cite}
\usepackage[dvipsnames]{xcolor}
\usepackage{amsmath,amssymb,amsfonts}
\usepackage{algorithmic}
\usepackage{graphicx}
\usepackage{textcomp}
\usepackage[dvipsnames]{xcolor}
\usepackage{amsmath,graphicx}
\usepackage{multirow}
\usepackage{makecell}
\usepackage{pgfplots}
\usepackage{adjustbox}
\usepackage{pgfplotstable}
\usepackage{standalone}
\usepackage{subcaption}
\usepackage{amssymb}
\usepackage{fixmath}
\usepackage[super]{nth}
\usetikzlibrary{patterns.meta}
\usepackage{tikz}
\usepackage{bm}
\usepackage{enumitem,kantlipsum}
\usepackage[flushleft]{threeparttable}

\def\BibTeX{{\rm B\kern-.05em{\sc i\kern-.025em b}\kern-.08em
    T\kern-.1667em\lower.7ex\hbox{E}\kern-.125emX}}

\pgfplotsset{
    my ybar legend/.style={
        legend image code/.code={
            \draw [##1] (0cm,-0.6ex) rectangle +(2.5em,1.5ex);
        },
        row sep=0.5cm
    },
}

\newcommand{\auxLoss}{Aux-Loss}
\newcommand{\simpleTopK}{\textit{Simple-Top-k}}

\newcommand{\zeroNorm}{$L_0$-Norm}
\newcommand{\indSoftmax}{\textit{OrthoSoftmax}}
\DeclareMathOperator{\E}{\mathbb{E}}
\newcommand{\LBS}{\textit{Librispeech}}
\newcommand{\ted}{\textit{TED-LIUM-v2}}
\newcommand{\LBSabb}{\textit{LBS}}
\newcommand{\tedabb}{\textit{TED-v2}}
\DeclareMathOperator{\softmax}{softmax}

\begin{document}

\title{Efficient Supernet Training with Orthogonal Softmax for Scalable ASR Model Compression\\

\thanks{\IEEEauthorrefmark{1} Denotes equal contribution}
}

\author{
    \IEEEauthorblockN{Jingjing Xu\IEEEauthorrefmark{1}\IEEEauthorrefmark{2}\IEEEauthorrefmark{3},
    Eugen Beck\IEEEauthorrefmark{1}\IEEEauthorrefmark{3},
    Zijian Yang\IEEEauthorrefmark{2}\IEEEauthorrefmark{3},
    Ralf Schlüter\IEEEauthorrefmark{2}\IEEEauthorrefmark{3}}
    \IEEEauthorblockA{\IEEEauthorrefmark{2}Machine Learning and Human Language
    Technology Group, RWTH Aachen University, Germany}
    \IEEEauthorblockA{\IEEEauthorrefmark{3}AppTek GmbH, Germany}
}


\maketitle

\begin{abstract}
  ASR systems are deployed across diverse environments, each with specific
  hardware constraints.
  We use supernet training to jointly train multiple encoders
  of varying sizes, enabling dynamic model size adjustment to fit hardware
  constraints without redundant training.
  Moreover, we introduce a novel method called {\indSoftmax}, which applies
  multiple orthogonal softmax functions to efficiently
  identify optimal subnets within the supernet,
  avoiding resource-intensive search.
  This approach also enables more flexible and precise subnet selection
  by allowing selection based on various criteria and levels of granularity.
  Our results with CTC on {\LBS} and {\ted} show that FLOPs-aware
  component-wise selection achieves the best overall performance.
  With the same number of training updates from one single job, WERs for all model
  sizes are comparable to or slightly better than those of individually trained models.
  Furthermore, we analyze patterns in the selected components and reveal interesting insights.
\end{abstract}


\section{Introduction}
\label{sec:intro}

Balancing accuracy and latency in automatic speech recognition (ASR) is a critical
 and active area of research.
The quest for higher accuracy has driven the development of larger, more complex
models  \cite{baevski2020wav2vec2,hsu2021hubert,radford23whisper}, resulting in
increased latency and higher energy consumption.
This trade-off is especially challenging in resource-constrained
environments like client-edge devices, where computational power
(measured in floating-point operations per second, or FLOPS) and memory
(the number of parameters) are limited \cite{gupta2024torque}.

To reduce the model size with minimal loss of performance, model
compression techniques such as pruning \cite{jiang2023microsoftpruning, peng2023l0norm, peng2023dphubert, yang2023l0norm},
knowledge distillation \cite{chang2022distillation,lee2022distillation,ashihara2022distillation},
quantization \cite{ding2024quantization,tybakov2023quantization}
are commonly used in ASR.
Quantization is orthogonal to other methods and can be applied to almost any
model.
However, the first two methods typically require a converged base model,
followed by separate fine-tuning for each compressed model of varying sizes,
causing redundant training and re-optimization efforts \cite{yuan2023metaomni2}.
Hence, another line of methods—supernet training,
 first proposed in \cite{yu2019sandwichrule}—aims to jointly train a supernet
encompassing multiple subnets with a single training process.
In this way, the supernet can generalize across various subnet configurations,
being adaptable to different model sizes.

Recent works \cite{ding2022cascaded,narayanan2021cascaded,shi2021dynamictransducer,yuan2023metaomni2,yang2022metaomni1,rui2022lighthubert}
in ASR have explored the concept of supernet training and achieved
promising results.
The challenge, however, remains in determining the optimal subnets efficiently
from the supernet.
The authors of \cite{ding2022cascaded,narayanan2021cascaded,shi2021dynamictransducer} identify
subnets by manually selecting the bottom layers or layers at regular intervals.
The authors of \cite{yuan2023metaomni2,yang2022metaomni1} use evolutionary search to find the
most performant subnets from a predefined search space at each training step.
Meanwhile, the authors of \cite{rui2022lighthubert,lee21pruning} conduct search after training.
In both cases, the search process is computationally expensive.
Recently, this issue is addressed in \cite{xu2024dynamic} by combining pruning
 with supernet training to automatically learn the optimal subnets during training.
However, the limitations are: 1) selection is restricted to entire layers, 2) the subnet size is determined by the number of layers,
without considering other criteria such as number of parameters,
3) the use of straight through estimator (STE) \cite{bengio2013ste} introduces gradient inconsistency, causing training instability and degraded performance  \cite{yin2019STE}.

Therefore, in this work, we propose a method called {\indSoftmax}, which apply
multiple softmax functions to independent, learnable score vectors.
We also introduce orthogonal constraint to let each softmax only select one
distinct group of parameters.
The softmax results are then aggregated to generate binary masks
that determine the subnets.
During training, this approach begins with a uniformly distributed
soft mask and then progressively converges to a binary hard mask for each subnet.
Compared to other commonly used binary mask learning approaches, this method avoids the
approximation inaccuracies commonly associated with continuous relaxations of
discrete operations, further details are discussed in Sec \ref{sec:jointlu_train_two_models}.
We also extend \cite{xu2024dynamic} by enabling finer selection
granularity and introducing additional selection criteria, which facilitates
more precise and flexible subnet selection.
We evaluate our approach by conducting experiments with the Conformer
\cite{gulati2020conformer} connectionist temporal classification (CTC)
\cite{graves2006ctc} model on both
{\LBS} and {\ted} datasets.
The results show that FLOPs-aware component-wise selection performs best, yielding
models that achieve WERs on par with individually
 trained ones while significantly reducing total training time.
We also analyze the patterns in component selection for subnets of varying sizes
in Sec \ref{sec:remain_ratio_analysis}.

\section{Efficient Supernet Training pipeline}
This section explains how we use supernet training to generate one supernet
and \(M\) subnets with fully shared parameters across varying sizes.
Let the acoustic encoder have learnable parameters \(\bm{\theta}\), decomposed
into \(N\) parameter groups \(\{\theta_j\}_{j=1}^N\).
The costs for each group are \(\bm{c} = \{c_j\}_{j=1}^N\).
Each of the \(M\) subnets has a different total cost constraint,
denoted as \(\{\tau_1, \tau_2, \ldots, \tau_M\}\).
For each subnetwork \( n_m \), we learn a binary mask \( \bm{z_m} \in \{0, 1\}^N \),
where parameter group \( \theta_j \) is retained in the encoder \(n_m\) if \( z_m^j = 1 \).
Therefore, we can formulate the joint training problem as:
\scalebox{0.82}{\parbox{1.2\linewidth}{%
\begin{flalign}
&\min\limits_{\theta, \{z_m\}_{m=1}^M} \E_{(x,y)\in \mathcal{D}} \left[ \mathcal{L}_{ASR}(x,y; \theta)\ + \sum_{m=1}^M \lambda_m \mathcal{L}_{ASR}(x,y; \theta \odot z_m)\right]\label{eq:joint_train_formulation}\\
&\text{subject to  } \sum_{j=1}^N z_m^j \cdot c_j < \tau_m, \forall m\in\{1, \ldots, M\},
\end{flalign}
}}\\
where \(\mathcal{D}\) is the training data, \(\odot\) denotes the element-wise product.
To better balance the loss scales between different encoders, we apply focal loss \cite{lin2017fl} as in
\cite{zhou2023enhancing} to compute \(\lambda_m\) (we set \(\beta_{focal}\) to 1),
where p denotes the model posterior of target sequence \(y\) given
input sequence \(x\):
\scalebox{0.88}{\parbox{1.1\linewidth}{%
\begin{align}
\lambda_m = [1-p(y|x, \theta \odot z_m)]^{\beta_{focal}} \label{eq:loss_scale_focal_loss}
\end{align}
}}

Our efficient training pipeline consists of two steps:
Step 1 aims at learning masks \(\bm{z_m}\) for each subnet \(n_m\).
Step 2 jointly trains the supernet and subnets determined by \(\bm{z_m}\)
to further enhance WER.
We allocate 60\% of the total training steps to Step 1 and 40\% to Step 2,
as recommended by  \cite{xu2024dynamic}.

\subsection{Step 1 - Learning Subnet Binary Mask}
To enable smooth adjustment and gradient-based optimization of component
inclusion, we propose \indSoftmax.
This method uses \(N\) softmax functions to compute a soft mask \(\bm{z_m} \in [0, 1]^N\),
where each value reflects the importance of different parameter groups.
We introduce a learnable score matrix \(\bm{S} \in \mathbb{R}^{N \times N}\),
initialized to zeros.
Applying \(N\) softmax functions to the rows of \(\bm{S}\) produces the weight
matrix \(\bm{W} \in {[0,1]}^{N\times N}\),
where
\(\bm{W}_{i,j} = {\softmax}_i (\bm{S}_{i,j})= \frac{e^{\bm{S}_{i,j}/T}}{{\sum_{j=1}^N e^{\bm{S}_{i,j}/T}}}\),
\(\softmax_i\) denotes the \(i^{th}\) softmax. Temperature \(T\) is annealed
from 1 to a minimum of 0.1
by factor of 0.999992 at each step.
Assuming the maximum number of parameter groups for subnetwork \(n_m\) that fit
the cost constraint \(\tau_m\) is \(k_m\),
we design the training to achieve the following objectives for the \(\bm{W}\):
\begin{enumerate}[leftmargin=*]
    \item Orthogonality: Ensure all row pairs in \(\bm{W}\) are orthogonal.
    This will force each softmax selects exactly one non-overlapping parameter group.
    \item Importance Ranking: Ensure that the parameter groups selected from the upper
    rows are more critical to ASR performance than those selected from the lower rows.
\end{enumerate}
we compute \(\bm{z_m}=\sum_{i=1}^{k_m} \bm{W}_{i,:}\),
\(\bm{z_m}\) initially starts with a uniform distribution and will naturally
converge to a \(k_m\)-hot vector during the training process. \\
\textbf{Determine \(k_m\) for Each Subnetwork \(n_m\).}
For each training step, we compute \(k_m\) that satisfies the cost constraint \(\tau_m\):
\scalebox{0.8}{\parbox{1.1\linewidth}{%
\begin{flalign}
\label{tab:equation_constraint}
k_m = \max \{k\} \text{  s.t.  } \sum_{i=1}^k \left( \sum_{j=1}^N \bm{W}_{ij} \cdot c_j \right) < \tau_m
\end{flalign}
}}\\
During training, \(k_m\) will adjust according to \(\bm{W}\).
As each row \(\bm{W}_{i,:}\) approaches a one-hot vector, \(k_m\) becomes
more reliable.\\
\textbf{Orthogonality.}
The orthogonal constraint helps achieve the design objective 1.
Orthogonal constraints have been used in other work \cite{zhang2021orthogonal,wang2020orthogonal} to improve the numerical
stability of matrices.
Assume \(k_m\) is computed in the current training step.
We apply the orthogonal constraint to the top \(k_m\) rows of the matrix \(\bm{W}\).
For all \(i, j \in \{1, \ldots, k_m\}\) with \(i \neq j\), we expect
\(\bm{W}_{i,:} \perp \bm{W}_{j.:}\), meaning \(\bm{W}_{i,:} \cdot \bm{W}_{j,:} = 0\).
Suppose the product of sub-matrix \(\bm{W}_{1:k_m,:}\) and its transpose
 \(\bm{W}_{1:k_m,:}^\intercal\) is \(\bm{D} \in \mathcal{R}^{k_m \times k_m}\).
We aim to force \(\bm{D}\) to be close to the
identity matrix \(\bm{I} \in \{0,1\}^{k_m \times k_m}\) by minimizing the
Frobenius norm of their difference.
Since \(\bm{D}\) is symmetric, we only take its upper triangle part,
\(\bm{D}_{\text{ut}} = \left[ \bm{D}_{ij} \mid i \leq j \right]\) for loss computation: \\
\scalebox{0.85}{\parbox{1.1\linewidth}{%
\begin{align}
  \mathcal{L}_{\text{orthog}} = \|\bm{D}_{\text{ut}} - \bm{I}\|_F = \sqrt{\sum_{i=1}^{k_m} (D_{i,i} - 1)^2 + \sum_{i=1}^{k_m} \sum_{j=i+1}^{k_m} D_{i,j}^2}
\end{align}
}}\\
\textbf{Importance Ranking.} Since we select rows from top to bottom, matrix
\(\bm{W}\) is forced to place parameter groups that are more critical for ASR
performance in upper rows.
To elaborate, \(\tau_m < \tau_{m'}\) implies \(k_m < k_{m'}\), the parameter groups
selected for subnet \(n_{m'}\) encompass those selected for subnet \(n_m\).
The \(k_m\) parameter groups for subnet \(n_{m}\) should be the most important
among \(k_{m'}\) parameter groups for subnet \(n_{m'}\).

\subsubsection{Training Loss}
The overall training loss of Step 1 can be formulated as: \\
\scalebox{0.82}{\parbox{1.2\linewidth}{%
\begin{flalign}
  \mathcal{L}_{ASR}(x,y; \theta)\ +  \lambda \mathcal{L}_{ASR}(x,y; \theta \odot z_m)
  + \beta \mathcal{L}_{orthog} (\bm{W}_{1:k_m, :}),
\label{eq:train_loss_step1}
\end{flalign}
}}\\
where \(m\) is randomly selected from the set \(\{1, \ldots, M\}\) at each
training step.
Compared to Eq.\ref{eq:joint_train_formulation}, Eq.\ref{eq:train_loss_step1}
trains only the supernet with one subnet to avoid scaling the training cost with
the number of subnets but still maintain overall performance.

\subsection{Step 2 - Supernet/Subnet Joint Train}
In Step 2, \(\bm{z_m}\) is rounded to an exact binary vector to define each \(n_m\).
We then jointly train the supernet with the selected subnets to further enhance performance.
To improve training efficiency, in each training step, we apply the sandwich
rules \cite{yu2019sandwichrule} by forwarding only three models at once: the smallest subnet, the largest
 subnet, and one medium subnet randomly sampled from the remaining \( M-2 \) subnets.
To mitigate mutual interference between the supernet and subnets, we apply layer dropout \cite{fan2020structuredpruning} on layers where all groups from that layer are not selected for the smallest subnet.
Additionally, we adapt the dropout magnitude in the feed-forward module (FFN) following \cite{yuan2023metaomni2}, scaling it by \( \frac{c_m}{c_s} \), where \( c_s\),
\( c_m\) is the number of hidden channels in the supernet and subnet \(n_m\) respectively.

\subsection{Selection Criteria and Granularity}
\label{sec:selection_criteria}
We decompose the Conformer encoder into distinct parameter groups with varying
granularities and assign each group a cost based on real-world requirements.\\
\textbf{Sparsity/FLOPs-aware Selection.}
We compute the cost of each parameter group using two metrics:
sparsity and the total number of floating-point operations (FLOPs).
Sparsity, which is the proportion of zero-valued parameters in a model,
directly affects model size and storage memory.
FLOPs is indicative of inference cost, making it crucial
for use cases where inference speed is critical.
We use fvcore\footnote{https://github.com/facebookresearch/fvcore}
 to calculate the FLOPs.\\
\textbf{Layer/Component-wise Selection.}
For layer-wise selection, we perform selection
directly on entire FFN, convolution module (Conv),
multi-headed self-attention module (MHSA) as in \cite{xu2024dynamic}.
Inspired by \cite{jiang2023microsoftpruning}, we extend the decomposition
granularity to components: hidden channel chunks in FFN,
attention heads in MHSA, and entire Conv layers.


%
%

\section{Experiments}
\label{sec:copyright}

\subsection{Experimental Setup}
Our experiments are conducted on 200h {\ted} ({\tedabb}) \cite{rousseau2014tedlium} and
960h {\LBS} ({\LBSabb}) \cite{vassil2015lbs960} dataset.
We use a phoneme-based CTC model from the previous work \cite{xu2024dynamic} as
baseline.
The output labels are 79 end-of-word augmented phonemes \cite{wei2021phoneme}.
The acoustic encoder consists of a VGG front end and 12 Conformer \cite{gulati2020conformer} blocks.
We set the model size $d_{model}$ to 512 for {\LBSabb} and 384 for {\tedabb} corpus, respectively.
The number of attention heads is $\frac{d_{model}}{64}$ and FFN
intermediate dimensions is $4 \times d_{model}$.
For component-wise selection in Sec \ref{sec:selection_criteria}, we set channel chunk size to \(d_{model}\), so each FFN layer is decomposed into 4 chunks.
The total number of parameters for {\tedabb} is 41.7M with 3.97$\times 10^8$ FLOPs, while for \text{\LBSabb} it is 74.2M with 6.19$\times 10^8$ FLOPs.
To ensure a fair comparison, we train for 730k steps in the {\LBSabb} experiments and for 380k steps in the {\tedabb} experiments across all methods.
The same learning rate schedule from \cite{xu2024dynamic} is used.
We apply Viterbi decoding with a 4-gram word-level language model.
We use RETURNN \cite{zeyer2018returnn} to train the acoustic models and RASR \cite{wiesler2014rasr} for recognition.
All our config files and code to reproduce the results can be found online\footnote{https://github.com/rwth-i6/returnn-experiments/tree/master/2024-orthogonal-softmax}.

\begin{table}[t]
  \centering
  \caption{\textit{ASR results comparison of different training methods, selection criteria, and granularities for training two models on the {\ted} test set.}}
  \label{tab:ted_jointly_train_two_model}
  \setlength{\tabcolsep}{0.1em}
  \scalebox{0.95}{\begin{tabular}{|l|c|c|c||c|c|c|}
  \hline
  \multirow{3}{*}{\shortstack{Training \\ method}} & \multirow{3}{*}{\shortstack{Select \\ criteria}} &
  \multirow{3}{*}{\shortstack{Select \\ granularity}} & \multicolumn{1}{c||}{Large} &
  \multicolumn{3}{c|}{Small} \\
  & & & \shortstack{WER \\ $[\%]$} & \shortstack{Params. \\ $[\text{M}]$} & \shortstack{FLOPs \\ $[10^8]$} &
  \shortstack{WER \\ $[\%]$} \\ \hline
  separately & \multirow{3}{*}{-} & - & 7.8 & \multirow{2}{*}{21.6} & \multirow{2}{*}{2.53} & 8.2 \\ \cline{1-1}\cline{3-4} \cline{7-7}
  \auxLoss &  & block & 7.7 & & & 8.8 \\ \cline{1-1}\cline{3-7}
  \multirow{5}{*}{\simpleTopK} & & \multirow{2}{*}{layer} & 7.7 & 20.7 & 2.47 & 8.3 \\ \cline{2-2} \cline{4-7}
  & \multirow{2}{*}{sparsity} &  & 7.7 & 21.3 & 2.45 & 8.2 \\ \cline{3-7}
  & & component & 7.8 & 21.0 & 2.49 & 8.5 \\ \cline{2-7}
  & \multirow{2}{*}{FLOPs} & layer & 7.7 & 20.9 & 2.46 & 8.2 \\ \cline{3-7}
  & & component & 7.9 & 20.4 & 2.45 & 8.6 \\ \hline

  \multirow{2}{*}{\zeroNorm} &   \multirow{2}{*}{sparsity} & layer & 8.0 & 20.8  & 2.48 & 8.2 \\ \cline{3-7}
   & & component & 8.1 & 20.7 & 2.48 & 8.5 \\ \hline

  \multirow{5}{*}{\indSoftmax} & - & \multirow{2}{*}{layer} & \textbf{7.6} & 22.8 & 2.62 & 8.3 \\ \cline{2-2} \cline{4-7}

  & \multirow{2}{*}{sparsity} & & 7.7 & 21.5 & 2.53 & 8.1 \\ \cline{3-7}
  & & component & 7.7 & 21.0 & 2.49 & 8.2 \\ \cline{2-7}

  & \multirow{2}{*}{FLOPs} & layer & 7.8 & 21.9 & 2.55 & 8.1  \\ \cline{3-7}
  & & component$^*$ & 7.7 & 21.5 & 2.52 & \textbf{8.0}  \\ \hline
\end{tabular}}\\
\leftline{{\scriptsize * used as the baseline for Table \ref{tab:loss_scale_cmp} and Table \ref{tab:training_trick_ablation}}}
\vspace{-2mm}
\end{table}

\begin{table}[t]
  \centering
  \caption{\textit{WERs[\%] for varying loss scales \(\lambda\) and \(\beta\) in training two models with {\indSoftmax} FLOPs-aware component-wise selection on the {\ted} dev and test set.}}
  \label{tab:loss_scale_cmp}
  \setlength{\tabcolsep}{0.9em}
  \scalebox{0.95}{\begin{tabular}{|c|c|c|c||c|c|}
  \hline
  \multirow{3}{*}{loss scale $\lambda$} & \multirow{3}{*}{loss scale $\beta$} & \multicolumn{4}{c|}{WER [\%]} \\ \cline{3-6}
  & & \multicolumn{2}{c|}{Large} & \multicolumn{2}{c|}{Small} \\

  & & dev & test & dev & test \\ \hline
   \multirow{4}{*}{adaptive} & 0.5 & 7.7 & 7.9 & \textbf{7.8} & 8.0 \\ \cline{2-6}
    & 1 & 7.7 & 7.8 & 7.9 & 8.2 \\ \cline{2-6}
    & 1.5 & 7.7 & 8.0 & 7.9 & 8.3 \\ \cline{2-6}
    & \multirow{4}{*}{0 $\rightarrow$ 1} & \textbf{7.5} & \textbf{7.7} & 7.9 & 8.0\\ \cline{1-1} \cline{3-6}
   0.1 & & \textbf{7.5} & 7.8 & 8.6 & 8.9\\ \cline{1-1} \cline{3-6}
   0.5 & & 7.6 & 7.9 & \textbf{7.8}& 8.3  \\ \cline{1-1} \cline{3-6}
   1 & & 7.7 & 7.9 & \textbf{7.8} & \textbf{7.9} \\ \hline

\end{tabular}}
\vspace{-3mm}
\end{table}

\begin{table}[t]
  \centering
  \caption{\textit{ASR results of ablation study on {\ted} test and dev set.
  Baseline is $^*$ in Table \ref{tab:ted_jointly_train_two_model}.}}
  \label{tab:training_trick_ablation}
  \setlength{\tabcolsep}{0.9em}
  \scalebox{1}{\begin{tabular}{|l|c|c||c|c|}
  \hline
  \multirow{3}{*}{Ablation Study} & \multicolumn{4}{c|}{WER [\%]} \\
  & \multicolumn{2}{c|}{Large} & \multicolumn{2}{c|}{Small} \\
  & dev & test & dev & test \\ \hline
  baseline & \textbf{7.5} & \textbf{7.7} & 7.9 & 8.0 \\ \hline
  - softmax temperature annealing & 7.6 & \textbf{7.7} & \textbf{7.7} & \textbf{7.9}\\ \hline
  - adaptive dropout in FFN & 7.6 & 7.9 & \textbf{7.7} & 8.1  \\ \hline
  - layer dropout & \textbf{7.5} & \textbf{7.7} & 7.8 & 8.0  \\ \hline
\end{tabular}}
\vspace{-3mm}
\end{table}

\subsection{1 Supernet + 1 Subnet}
\label{sec:jointlu_train_two_models}
Table \ref{tab:ted_jointly_train_two_model} compares ASR results using
different training methods, selection criteria, and granularities on a full-size
supernet and a subnet about half its size.
``Separately'' refers to training and tuning the models individually
for each size category.
For the {\auxLoss}, we add intermediate CTC loss \cite{lee2021intermediatectc}
to the output of 6-th Conformer block.
We extend {\simpleTopK} \cite{xu2024dynamic} by applying selection criteria.
We also compare with {\zeroNorm} \cite{louizos2017l0nrom}, which is widely used in pruning ASR models \cite{jiang2023microsoftpruning,
 peng2023l0norm, yang2023l0norm, peng2023dphubert} to identify the subnet.
We integrate {\zeroNorm} into Step 1 to calculate the mask \(\bm{z}\).
We use the implementation from \cite{xia2022l0norm} and keep all other training aspects unchanged for a fair comparison.
The results show that the WERs of {\simpleTopK} and {\zeroNorm} degrade for both large and small models as select granularity becomes finer.
This contrasts with the {\indSoftmax} results, where finer granularity improves subnet selection.
Such behavior stems from inherent limitations in both methods.
{\simpleTopK} uses STE and suffers from gradient inconsistency.
{\zeroNorm} uses a hard concrete distribution to relax the binary mask \(\bm{z}\) with a random variable
\( u \sim U(0,1) \).
While this makes \(\bm{z}\) differentiable, it also introduces
 training instability due to the fluctuating influence of \(u\).
Finer granularity increases binary elements in \(\bm{z}\) to be relaxed, exacerbating gradient approximation errors and hindering model convergence.
In contrast, {\indSoftmax} updates \(\bm{z}\) smoothly across each training step.

\subsection{Ablation Study}
\label{sec:ablation_study}

Table \ref{tab:loss_scale_cmp} compares WERs across different loss scales.
The term ``adaptive'' refers to using
Eq.\ref{eq:loss_scale_focal_loss}, to dynamically adjust the subnet loss scale
at each training step based on its CTC probability.
The results show that compared to constant values, the adaptive approach
 effectively balances the supernet and subnet, resulting in strong performance
  for both.
The term “\(0\rightarrow1\)” indicates that \(\beta\) is linearly increased
from 0 to 1 during Step 1.
A larger \(\beta\) worsens WER for smaller encoders, likely due to premature
 subnet selection from accelerated softmax orthogonalization.
Contrarily, “\(0\rightarrow1\)” prioritizes ASR loss early, enabling
more informed component selection later when their impact on ASR performance
is clearer.


%

Table \ref{tab:training_trick_ablation} shows the impact of each training method.
Removing softmax \(T\) annealing in Step 1 slightly improves WER for small encoders.
However, empirical observations show that for larger \( k \),  \(T\) annealing
can accelerate softmax orthogonalization and help convergence.
Therefore, we retain it in our approach.


\subsection{1 Supernet + 2 Subnets}
\label{sec:1_supernet_2_subnet}
Table \ref{tab:tedlium_three_model} and Table \ref{tab:lbs_three_model} report
the results of jointly training
three encoders on {\tedabb} and {\LBSabb} test set.
{\indSoftmax} significantly outperforms {\auxLoss} on medium and
small models.
WER discrepancies increase as model size decreases, likely
because {\auxLoss} leverages the lower layers at the bottom, which are used to
capture low-level features.
For {\indSoftmax}, FLOPs-aware selection generally performs better than
sparsity-aware selection.
Using components as the granularity of selection proves more effective than using
layers, especially for small-sized models.
The WERs with FLOPs-aware component-wise selection are at least as good as those
of the individually trained model across all model sizes and datasets, while
significantly reducing training time.

\begin{table}[t!]
  \centering
  \caption{\textit{ASR results of three encoders of large, medium, and small
  sizes on {\ted} test set}}
  \label{tab:tedlium_three_model}
  \setlength{\tabcolsep}{0.1em}
  \scalebox{0.95}{\begin{tabular}{|l||c|c|c||c|c|c||c|c|c|}
  \hline
  \multirow{2}{*}{Training} &  \multirow{2}{*}{\shortstack{Select \\ criteria}} &
  \multirow{2}{*}{\shortstack{Select \\ granularity}} & Large & \multicolumn{3}{c||}{Medium} &
  \multicolumn{3}{c|}{Small} \\
  & & & \shortstack{WER\\$[\%]$} & \shortstack{Parm.\\$[\text{M}]$}
  & \shortstack{FLOPs \\ $[10^8]$} & \shortstack{WER\\$[\%]$} & \shortstack{Parm.\\$[\text{M}]$}
  & \shortstack{FLOPs \\ $[10^8]$} & \shortstack{WER\\$[\%]$} \\ \hline

  separately & \multirow{2}{*}{-} & - & 7.8 & \multirow{2}{*}{28.4} & \multirow{2}{*}{3.01} & 8.1 & \multirow{2}{*}{14.7} & \multirow{2}{*}{2.06} & 8.9\\ \cline{1-1}\cline{3-4}\cline{7-7}\cline{10-10}
  \auxLoss & & block & \textbf{7.6} & & & 8.0 & & & 10.1 \\ \hline

  \multirow{4}{*}{\shortstack{\textit{Ortho-} \\ \textit{Softmax}}} &
  \multirow{2}{*}{sparsity} & layer & 7.7 & 27.7 & 2.96 & 8.0 &
  14.9 & 2.06 & 9.0 \\ \cline{3-10}
   & & component & 7.8 & 28.1 & 2.99 & 8.1 & 14.8 & 2.06 & 8.8 \\ \cline{2-10}

   & \multirow{2}{*}{FLOPs} & layer & 7.7 & 28.3 & 3.00 & \textbf{7.9} & 15.3 & 2.09 & 8.7 \\ \cline{3-10}
   & & component & 7.8 & 28.3 & 3.00 & 8.0 & 15.0 & 2.07 & \textbf{8.6}  \\ \hline

\end{tabular}}
\end{table}

\begin{table}[t!]
  \centering
  \caption{\textit{ASR results of three encoders of large, medium, and small
  sizes on {\LBS} test-clean and test-other set.}}
  \label{tab:lbs_three_model}
  \setlength{\tabcolsep}{0.05em}
  \scalebox{0.95}{\begin{tabular}{|l||c|c|c|c||c|c|c|c||c|c|c|c|}
  \hline
  \multirow{3}{*}{\shortstack{Training \\ method}} &  \multirow{3}{*}{\shortstack{Select \\ criteria}} &
  \multirow{3}{*}{\shortstack{Select \\ granl.}} & \multicolumn{2}{c||}{Large} & \multicolumn{4}{c||}{Medium} &
  \multicolumn{4}{c|}{Small} \\
  & & & \multicolumn{2}{c||}{\shortstack{WER$[\%]$}} & \multirow{2}{*}{\shortstack{Prm.\\$[\text{M}]$}}
  & \multirow{2}{*}{\shortstack{FLOPs \\ $[10^8]$}} & \multicolumn{2}{c||}{\shortstack{WER$[\%]$}} & \multirow{2}{*}{\shortstack{Prm.\\$[\text{M}]$}}
  & \multirow{2}{*}{\shortstack{FLOPs \\ $[10^8]$}} & \multicolumn{2}{c|}{\shortstack{WER$[\%]$}} \\
  & & & cln & oth & & & cln & oth & & & cln & oth \\ \hline

  separately & \multirow{2}{*}{-} & - & 3.3 & 7.1 & \multirow{2}{*}{49.9} &
  \multirow{2}{*}{4.50} & 3.5 & 7.7 & \multirow{2}{*}{25.7} &
  \multirow{2}{*}{2.80} & \textbf{3.6} & \textbf{8.4} \\ \cline{1-1} \cline{3-5} \cline{8-9} \cline{12-13}
  \auxLoss & & block & \textbf{3.2} & \textbf{6.9} & & & 3.6 & 7.9 & & & 4.5 & 9.7  \\
  \cline{1-1} \cline{2-13}

  \multirow{4}{*}{\shortstack{\textit{Ortho-} \\ \textit{Softmax}}} &
  \multirow{2}{*}{sparsity} & layer & \textbf{3.2} & 7.1 & 49.8 & 4.48 & 3.3 & 7.4 & 25.0
  & 2.74 & 3.8 & 9.0 \\ \cline{3-13}
   & & cmp & \textbf{3.2} & \textbf{6.9} & 49.8 & 4.48 & 3.3 & \textbf{7.3} & 24.4 & 2.71 & 3.7 & 8.8  \\ \cline{2-13}

   & \multirow{2}{*}{FLOPs} & layer & \textbf{3.2} & 7.1 & 48.7 & 4.41 & \textbf{3.2} & 7.4 & 26.1
   & 2.81 & 3.7 & 8.7 \\ \cline{3-13}
   & & cmp & \textbf{3.2} & 7.0 & 49.8 & 4.48 & 3.4 & 7.5 & 25.6 & 2.79 & \textbf{3.6} & \textbf{8.4} \\ \hline
\end{tabular}}
\vspace{-3mm}
\end{table}

\subsection{1 Supernet + 4 Subnets}
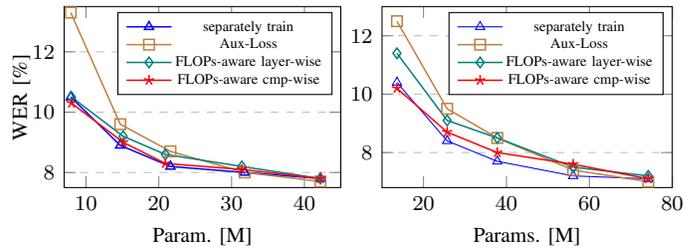
\begin{figure}
     \centering
     \subfloat[][{\ted} test]{\input{tikzpicture/ted_five_models.tex}}
     \subfloat[][{\LBS} test-other]{\input{tikzpicture/lbs_five_models.tex}}
     \caption{ASR results comparison of training five encoders. 
     }
     \label{fig:joint_train_five_models}
     \vspace{-2mm}
\end{figure}

The results shown in Fig.\ref{fig:joint_train_five_models} are consistent with the observations in Sec \ref{sec:1_supernet_2_subnet}.
For both {\tedabb} and {\LBSabb} test sets, the red line closely matches the blue line, indicating that the FLOPs-aware component-wise approach achieves on-par WER across all five models compared to individual training.


\subsection{Remaining Ratio Analysis}
\label{sec:remain_ratio_analysis}
\begin{figure}[t!]
  \captionsetup[subfloat]{position=bottom,labelformat=empty, belowskip=-1pt,
  aboveskip=-1pt, font=scriptsize}
  \centering
  \subfloat[1.94 $\times 10^8$ (34.1\%) FLOPs]{
  \input{tikzpicture/model_1.tex}
  }
  \vspace{0mm}
  \subfloat[2.79 $\times 10^8$ (45.1\%) FLOPs]{
  \input{tikzpicture/model_2.tex}
  }
  \vspace{0mm}
  \subfloat[3.65 $\times 10^8$ (58.8\%) FLOPs]{
  \input{tikzpicture/model_3.tex}
  }
  \vspace{0mm}
  \subfloat[4.92 $\times 10^8$ (79.3\%) FLOPs]{
  \input{tikzpicture/model_4.tex}
  }
  \caption{The remaining ratio for each layer in encoders
  of varying sizes, trained using the FLOPs-aware component-wise {\indSoftmax}
  method, as shown in Fig.\ref{fig:joint_train_five_models}b.}
  \label{fig:remain_ratio}
  \vspace{-4mm}
\end{figure}
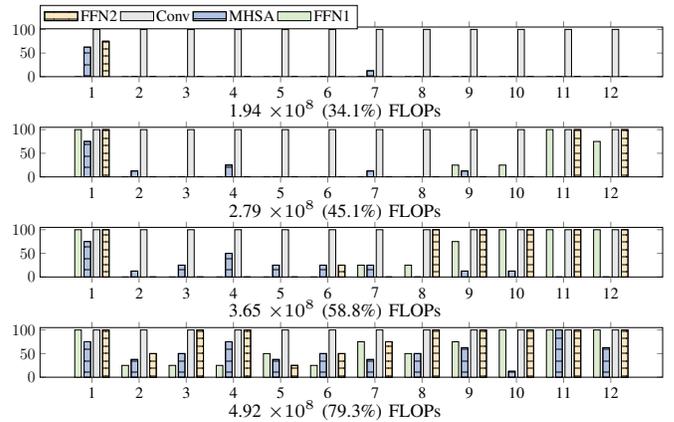

Fig.\ref{fig:remain_ratio} shows that Convs are retained the
most in all models, which is also observed in
\cite{xu2024dynamic,jiang2023microsoftpruning}, indicating that Convs are
generally more important for the Conformer model.
Besides, we observe that a large portion of the $1^{st}$ block is
retained, even in the smallest model.
This is likely because the $1^{st}$ block captures the initial temporal and
phonetic characteristics.
Also, fewer parameters are preserved in the lower layers compared to
the higher layers, suggesting that the lower layers of the Conformer contain
more redundancy.
Furthermore, the MHSA heads are distributed sparsely across blocks.
The authors of \cite{voita2019att} find that only a small subset of heads is crucial, and
 removing the vast majority of heads does not significantly affect performance.
For smaller models, no MHSAs are preserved in the upper two layers, which
aligns with the conclusion from \cite{zhang2021att} that these upper attention
layers may be less useful due to the nearly diagonal attention score matrix.
Although not shown in Fig.\ref{fig:remain_ratio}, we observe that sparsity-aware
approaches also select Convs the most.
However, compared to the FLOPs-aware method, they favor FFN components
over MHSA heads.

\section{Conclusion}
In this work, we introduce {\indSoftmax} for learning binary masks to
 identify optimal subnets during supernet training.
Additionally, our approach enables subnet selection based on different levels of
granularity and selection criteria.
Experimental results show that FLOPs-aware component-wise selection performs
overall the best.
With the same number of training updates as a single training job, this approach
achieves WERs comparable to or even better than those of
individually trained models across various model sizes.
By analyzing the selected components, we find that the 1$^{st}$ Conformer block
is important, and Convs are the most critical component.

\section*{Acknowledgement}
\scriptsize
This work was partially supported by NeuroSys, which as part of the
initiative “Clusters4Future” is funded by the Federal Ministry of
Education and Research BMBF (03ZU1106DA), and by the project RESCALE
within the program \textit{AI Lighthouse Projects for the Environment,
Climate, Nature and Resources} funded by the Federal Ministry for the
Environment, Nature Conservation, Nuclear Safety and Consumer
Protection (BMUV), funding ID: 6KI32006B.

\end{document}

%% file: tikzpicture/ted_five_models.tex
\begin{tikzpicture}
\begin{axis}[
    width=0.29\textwidth,
    height=0.22\textwidth,
    ylabel={WER [\%]},
    y label style={at={(0.18,0.5)}},
    xlabel={Param. [M]},
    x label style={at={(0.5,0.05)}},
    xmin=7, xmax=45,
    ymin=7.5, ymax=13.5,
    legend style={nodes={scale=0.55}},
    legend pos=north east,
    ymajorgrids=true,
    grid style=dashed,
    tick label style={font=\footnotesize},
    xlabel style={font=\footnotesize},
    ylabel style={font=\footnotesize}
]

\addplot[line width=0.2mm,color=blue,mark=triangle,]
    coordinates {
    (7.9, 10.5) 
    (14.7, 8.9) 
    (21.6, 8.2) 
    (31.8, 8.0) 
    (42.2, 7.8) 
    };

\addplot[line width=0.2mm,color=brown,mark=square,]
    coordinates {
    (7.9, 13.3) 
    (14.7, 9.6) 
    (21.6, 8.7) 
    (31.8, 8.0) 
    (42.2, 7.7) 
    };

\addplot[line width=0.2mm,color=teal,mark=diamond,]
    coordinates {
    (8.0, 10.5)
    (15.1, 9.2)
    (20.9, 8.6)
    (31.4, 8.2)
    (42.2, 7.8)
    };

\addplot[line width=0.2mm,color=red,mark=star,]
    coordinates {
    (8.0, 10.3)
    (15.1, 9.0)
    (20.9, 8.3)
    (31.4, 8.1)
    (42.2, 7.8)
    };

\legend{separately train, \auxLoss, FLOPs-aware layer-wise, FLOPs-aware cmp-wise}
\end{axis}
\end{tikzpicture}

%% file: tikzpicture/lbs_five_models.tex
\begin{tikzpicture}
\begin{axis}[
    width=0.3\textwidth,
    height=0.22\textwidth,
    xlabel={Params. [M]},
        x label style={at={(0.5,0.05)}},
    xmin=10, xmax=80,
    ymin=6.8, ymax=13,
    legend style={nodes={scale=0.55}},
    legend pos=north east,
    ymajorgrids=true,
    grid style=dashed,
    tick label style={font=\footnotesize},
    xlabel style={font=\footnotesize}
]

\addplot[color=blue,mark=triangle,]
    coordinates {
      (13.6, 10.4)
      (25.7, 8.4)
      (37.8, 7.7)
      (56.1, 7.2)
      (74.2, 7.1)
    };

  \addplot[line width=0.2mm,color=brown,mark=square,]
      coordinates {
      (13.6, 12.5)
      (25.7, 9.5)
      (37.8, 8.5)
      (56.1, 7.4)
      (74.2, 7.0)
      };

\addplot[line width=0.2mm,color=teal,mark=diamond,]
    coordinates {
      (13.6, 11.4)
      (25.7, 9.1)
      (37.8, 8.5)
      (56.1, 7.5)
      (74.2, 7.2)
    };

\addplot[line width=0.2mm,color=red,mark=star,]
    coordinates {
      (13.6, 10.2)
      (25.7, 8.7)
      (37.8, 8.0)
      (56.1, 7.6)
      (74.2, 7.1)
    };

\legend{separately train, \auxLoss, FLOPs-aware layer-wise, FLOPs-aware cmp-wise}
\end{axis}
\end{tikzpicture}

%% file: tikzpicture/model_1.tex
\pgfplotstableread{
Label FFN1 Conv MHSA FFN2
1 0 100 62.5 75
2 0 100 0 0
3 0 100 0 0
4 0 100 0 0
5 0 100 0 0
6 0 100 0 0
7 0 100 12.5 0
8 0 100 0 0
9 0 100 0 0
10 0 100 0 0
11 0 100 0 0
12 0 100 0 0
}\data

\begin{tikzpicture}[scale=0.5]
  \tikzstyle{every node}=[font=\large]
    \begin{axis}[
        ybar,
        bar width=5,
        width=\textwidth,
        height=.16\textwidth,
        ymin=0,
        ymax=105,
        xtick=data,
        legend columns=4,
        legend style={at={(0,1.2)},anchor=west, row sep=0.5cm},
        reverse legend=true,
        xticklabels from table={\data}{Label},
        xticklabel style={text width=3cm,align=center},
        my ybar legend,
    ]
        \addplot [fill=YellowGreen!30]
            table [y=FFN1, meta=Label, x expr=\coordindex]
                {\data};
                    \addlegendentry{FFN1}
        \addplot [fill=NavyBlue!30!, postaction={pattern={Lines[angle=0,
        distance=1.5mm, line width=0.03mm]}}]
            table [y=MHSA, meta=Label, x expr=\coordindex]
                {\data};
                    \addlegendentry{MHSA}

        \addplot [fill=Gray!20]
            table [y=Conv, meta=Label, x expr=\coordindex]
                {\data};
                    \addlegendentry{Conv}
        \addplot [fill=Dandelion!30,postaction={pattern={Lines[angle=45,
        distance=1mm, line width=0.03mm]}},]
            table [y=FFN2, meta=Label, x expr=\coordindex]
                {\data};
                \addlegendentry{FFN2}
    \end{axis}
\end{tikzpicture}

%% file: tikzpicture/model_2.tex
\pgfplotstableread{
Label FFN1 Conv MHSA FFN2
1 100 100 75 100
2 0 100 12.5 0
3 0 100 0 0
4 0 100 25 0
5 0 100 0 0
6 0 100 0 0
7 0 100 12.5 0
8 0 100 0 0
9 25 100 12.5 0
10 25 100 0 0
11 100 100 0 100
12 75 100 0 100
}\data

\begin{tikzpicture}[scale=0.5]
  \tikzstyle{every node}=[font=\large]
    \begin{axis}[
        ybar,
        bar width=5,
        width=\textwidth,
        height=.16\textwidth,
        ymin=0,
        ymax=105,
        xtick=data,
        legend columns=-1,
        legend style={at={(0.5,1.2)},anchor=west},
        reverse legend=true,
        xticklabels from table={\data}{Label},
        xticklabel style={text width=3cm,align=center},
    ]
        \addplot [fill=YellowGreen!30]
            table [y=FFN1, meta=Label, x expr=\coordindex]
                {\data};
        \addplot [fill=NavyBlue!30!, postaction={pattern={Lines[angle=0,
        distance=1.5mm, line width=0.03mm]}}]
            table [y=MHSA, meta=Label, x expr=\coordindex]
                {\data};
        \addplot [fill=Gray!20]
            table [y=Conv, meta=Label, x expr=\coordindex]
                {\data};
        \addplot [fill=Dandelion!30,postaction={pattern={Lines[angle=45,
        distance=1mm, line width=0.03mm]}},]
            table [y=FFN2, meta=Label, x expr=\coordindex]
                {\data};

    \end{axis}
\end{tikzpicture}

%% file: tikzpicture/model_3.tex
\pgfplotstableread{
Label FFN1 Conv MHSA FFN2
1 100 100 75 100
2 0 100 12.5 0
3 0 100 25 0
4 0 100 50 0
5 0 100 25 0
6 0 100 25 25
7 25 100 25 0
8 25 100 0 100
9 75 100 12.5 100
10 100 100 12.5 100
11 100 100 0 100
12 100 100 0 100
}\data

\begin{tikzpicture}[scale=0.5]
  \tikzstyle{every node}=[font=\large]
    \begin{axis}[
        ybar,
        bar width=5,
        width=\textwidth,
        height=.16\textwidth,
        ymin=0,
        ymax=105,
        xtick=data,
        legend columns=-1,
        legend style={at={(0.5,1.2)},anchor=west},
        reverse legend=true,
        xticklabels from table={\data}{Label},
        xticklabel style={text width=3cm,align=center},
    ]
        \addplot [fill=YellowGreen!30]
            table [y=FFN1, meta=Label, x expr=\coordindex]
                {\data};
        \addplot [fill=NavyBlue!30!, postaction={pattern={Lines[angle=0,
        distance=1.5mm, line width=0.03mm]}}]
            table [y=MHSA, meta=Label, x expr=\coordindex]
                {\data};
        \addplot [fill=Gray!20]
            table [y=Conv, meta=Label, x expr=\coordindex]
                {\data};
        \addplot [fill=Dandelion!30,postaction={pattern={Lines[angle=45,
        distance=1mm, line width=0.03mm]}},]
            table [y=FFN2, meta=Label, x expr=\coordindex]
                {\data};

    \end{axis}
\end{tikzpicture}

%% file: tikzpicture/model_4.tex
\pgfplotstableread{
Label FFN1 Conv MHSA FFN2
1 100 100 75 100
2 25 100 37.5 50
3 25 100 50 100
4 25 100 75 100
5 50 100 37.5 25
6 25 100 50 50
7 75 100 37.5 75
8 50 100 50 100
9 75 100 62.5 100
10 100 100 12.5 100
11 100 100 100 100
12 100 100 62.5 100
}\data

\begin{tikzpicture}[scale=0.5]
  \tikzstyle{every node}=[font=\large]
    \begin{axis}[
        ybar,
        bar width=5,
        width=\textwidth,
        height=.16\textwidth,
        ymin=0,
        ymax=105,
        xtick=data,
        legend columns=-1,
        legend style={at={(0.5,1.2)},anchor=west},
        reverse legend=true,
        xticklabels from table={\data}{Label},
        xticklabel style={text width=3cm,align=center},
    ]
        \addplot [fill=YellowGreen!30]
            table [y=FFN1, meta=Label, x expr=\coordindex]
                {\data};
        \addplot [fill=NavyBlue!30!, postaction={pattern={Lines[angle=0,
        distance=1.5mm, line width=0.03mm]}}]
            table [y=MHSA, meta=Label, x expr=\coordindex]
                {\data};
        \addplot [fill=Gray!20]
            table [y=Conv, meta=Label, x expr=\coordindex]
                {\data};
        \addplot [fill=Dandelion!30,postaction={pattern={Lines[angle=45,
        distance=1mm, line width=0.03mm]}},]
            table [y=FFN2, meta=Label, x expr=\coordindex]
                {\data};

    \end{axis}
\end{tikzpicture}

%% file: IEEE-conference-template-062824.bbl
\begin{thebibliography}{00}
\bibitem{baevski2020wav2vec2}
A.~Baevski, Y.~Zhou, A.~Mohamed, and M.~Auli,
\newblock ``{wav2vec 2.0: {A} Framework for Self-Supervised Learning of Speech
  Representations},''
\newblock in {\em NeurIPS}, virtual, Dec. 2020.

\bibitem{hsu2021hubert}
W.~Hsu, B.~Bolte, Y.~H. Tsai, K.~Lakhotia, R.~Salakhutdinov, and A.~Mohamed,
\newblock ``{HuBERT: Self-Supervised Speech Representation Learning by Masked
  Prediction of Hidden Units},''
\newblock {\em {IEEE} {ACM} Trans. Audio Speech Lang. Process.}, vol. 29, pp.
  3451--3460, 2021.

\bibitem{radford23whisper}
A.~Radford, J.~W. Kim, T.~Xu, G.~Brockman, C.~McLeavey, and I.~Sutskever,
\newblock ``{Robust Speech Recognition via Large-Scale Weak Supervision},''
\newblock in {\em {ICML}}, Hawaii, USA, July 2023, pp. 28492--28518.

\bibitem{gupta2024torque}
A.~Gupta, T.~Bau, J.~Kim, Z.~Zhu, S.~Jha, and H.~Garud,
\newblock ``{Torque based Structured Pruning for Deep Neural Network},''
\newblock in {\em {WACV}}, Waikoloa, HI, USA, Jan. 2024, pp. 2699--2708.

\bibitem{jiang2023microsoftpruning}
H.~Jiang, L.~L. Zhang, Y.~Li, Y.~Wu, S.~Cao, T.~Cao, Y.~Yang, J.~Li, M.~Yang,
  and L.~Qiu,
\newblock ``{Accurate and Structured Pruning for Efficient Automatic Speech
  Recognition},''
\newblock in {\em Interspeech}, Dublin, Ireland, Aug. 2023, pp. 4104--4108.

\bibitem{peng2023l0norm}
Y.~Peng, K.~Kim, F.~Wu, P.~Sridhar, and S.~Watanabe,
\newblock ``{Structured Pruning of Self-Supervised Pre-Trained Models for
  Speech Recognition and Understanding},''
\newblock in {\em {ICASSP}}, Greece, June 2023, pp. 1--5.

\bibitem{peng2023dphubert}
Y.~Peng, Y.~Sudo, M.~Shakeel, and S.~Watanabe,
\newblock ``{DPHuBERT: Joint Distillation and Pruning of Self-Supervised Speech
                  Models},''
\newblock in {\em Interspeech}, Dublin, Ireland, Aug. 2023, pp. 62--66.

\bibitem{yang2023l0norm}
H.~Wang, S.~Wang, W.~Zhang, H.~Suo, and Y.~Wan,
\newblock ``{Task-Agnostic Structured Pruning of Speech Representation
  Models},''
\newblock in {\em Interspeech}, Dublin, Ireland, Aug. 2023, pp. 4104--4108.

\bibitem{chang2022distillation}
H.~Chang, S.~Yang, and H.~Lee,
\newblock ``{Distilhubert: Speech Representation Learning by Layer-Wise
  Distillation of Hidden-Unit Bert},''
\newblock in {\em {ICASSP}}, Singapore, May 2022, pp. 7087--7091.

\bibitem{lee2022distillation}
Y.~Lee, K.~Jang, J.~Goo, Y.~Jung, and H.~R. Kim,
\newblock ``{FitHuBERT: Going Thinner and Deeper for Knowledge Distillation of
  Speech Self-Supervised Models},''
\newblock in {\em Interspeech}, Incheon, Korea, Sept. 2022, pp. 3588--3592.

\bibitem{ashihara2022distillation}
T.~Ashihara, T.~Moriya, K.~Matsuura, and T.~Tanaka,
\newblock ``{Deep versus Wide: An Analysis of Student Architectures for
  Task-Agnostic Knowledge Distillation of Self-Supervised Speech Models},''
\newblock in {\em Interspeech}, Incheon, Korea, Sept. 2022, pp. 411--415.

\bibitem{ding2024quantization}
S.~Ding, D.~Qiu, D.~Rim, Y.~He, O.~Rybakov, B.~Li, R.~Prabhavalkar, W.~Wang,
  T.~N. Sainath, Z.~Han, J.~Li, A.~Yazdanbakhsh, and S.~Agrawal,
\newblock ``{USM-Lite: Quantization and Sparsity Aware Fine-Tuning for Speech
  Recognition with Universal Speech Models},''
\newblock in {\em {ICASSP}}, Seoul, Korea, Apr. 2024, pp. 10756--10760.

\bibitem{tybakov2023quantization}
O.~Rybakov, P.~Meadowlark, S.~Ding, D.~Qiu, J.~Li, D.~Rim, and Y.~He,
\newblock ``{2-bit Conformer Quantization for Automatic Speech Recognition},''
\newblock in {\em Interspeech}, Dublin, Ireland, Aug. 2023, pp. 4908--4912.

\bibitem{yuan2023metaomni2}
Y.~Shangguan, H.~Yang, D.~Li, C.~Wu, Y.~Fathullah, D.~Wang, A.~Dalmia,
  R.~Krishnamoorthi, O.~Kalinli, J.~Jia, J.~Mahadeokar, X.~Lei, M.~Seltzer, and
  V.~Chandra,
\newblock ``{TODM: Train Once Deploy Many Efficient Supernet-Based RNN-T
  Compression For On-device ASR Models},''
\newblock in {\em {ICASSP}}, Seoul, Korea, Apr. 2024, pp. 10216--10220.

\bibitem{yu2019sandwichrule}
J.~Yu, L.~Yang, N.~Xu, J.~Yang, and T.~S. Huang,
\newblock ``{Slimmable Neural Networks},''
\newblock in {\em {ICLR}}, New Orleans, USA, May 2019.

\bibitem{ding2022cascaded}
S.~Ding, W.~Wang, D.~Zhao, T.~N. Sainath, Y.~He, R.~David, R.~Botros, X.~Wang,
  R.~Panigrahy, Q.~Liang, D.~Hwang, I.~McGraw, R.~Prabhavalkar, and
  T.~Strohman,
\newblock ``{A Unified Cascaded Encoder {ASR} Model for Dynamic Model Sizes},''
\newblock in {\em Interspeech}, Incheon, Korea, Sept. 2022, pp. 1706--1710.

\bibitem{narayanan2021cascaded}
A.~Narayanan, T.~N. Sainath, R.~Pang, J.~Yu, C.~Chiu, R.~Prabhavalkar,
  E.~Variani, and T.~Strohman,
\newblock ``{Cascaded Encoders for Unifying Streaming and Non-Streaming
  {ASR}},''
\newblock in {\em {ICASSP}}, Toronto, Canada, June 2021, pp. 5629--5633.

\bibitem{shi2021dynamictransducer}
Y.~Shi, V.~Nagaraja, C.~Wu, J.~Mahadeokar, D.~Le, R.~Prabhavalkar, A.~Xiao,
  C.~Yeh, J.~Chan, C.~Fuegen, O.~Kalinli, and M.~L. Seltzer,
\newblock ``{Dynamic Encoder Transducer: {A} Flexible Solution for Trading Off
  Accuracy for Latency},''
\newblock in {\em Interspeech}, Brno, Czechia, Aug. 2021, pp. 2042--2046.

\bibitem{yang2022metaomni1}
H.~Yang, S.~Yuan, D.~Wang, M.~Li, P.~Chuang, X.~Zhang, G.~Venkatesh,
O.~Kalinli, and V.~Chandra,
\newblock ``{Omni-Sparsity {DNN:} Fast Sparsity Optimization for On-Device
  Streaming {E2E} {ASR} Via Supernet},''
\newblock in {\em {ICASSP}}, Virtual and Singapore, May 2022, pp. 8197--8201.

\bibitem{rui2022lighthubert}
R.~Wang, Q.~Bai, J.~Ao, L.~Zhou, Z.~Xiong, Z.~Wei, Y.~Zhang, T.~Ko, and H.~Li,
\newblock ``{Lighthubert: Lightweight and Configurable Speech Representation
  Learning with Once-for-all Hidden-unit Bert},''
\newblock in {\em Interspeech}, Incheon, Korea, Sept. 2022, pp. 1686--1690.

\bibitem{lee21pruning}
J.~Lee, J.~Kang, and S.~Watanabe,
\newblock ``{Layer Pruning on Demand with Intermediate {CTC}},''
\newblock in {\em Interspeech}, Brno, Czechia.

\bibitem{xu2024dynamic}
J.~Xu, W.~Zhou, Z.~Yang, E.~Beck, and R.~Schlüter,
\newblock ``{Dynamic Encoder Size Based on Data-Driven Layer-wise Pruning for
  Speech Recognition},''
\newblock in {\em Interspeech}, Kos, Greece, Sept. 2024,
\newblock To Appear.

\bibitem{bengio2013ste}
Y.~Bengio, N.~Leonard, and A.~Courville,
\newblock ``{Estimating or Propagating Gradients Through Stochastic Neurons for Conditional Computation},''
\newblock arXiv:1308.3432, Aug.2013.

\bibitem{yin2019STE}
P.~Yin, J.~Lyu, S.~Zhang, S.~J. Osher, Y.~Qi, and J.~Xin,
\newblock ``{Understanding Straight-Through Estimator in Training Activation
  Quantized Neural Nets},''
\newblock in {\em {ICLR}}, New Orleans, LA, USA, May 2019.


\bibitem{gulati2020conformer}
A.~Gulati, J.~Qin, C.~Chiu, N.~Parmar, Y.~Zhang, J.~Yu, W.~Han, S.~Wang,
  Z.~Zhang, Y.~Wu, and R.~Pang,
\newblock ``{Conformer: Convolution-augmented Transformer for Speech
  Recognition},''
\newblock in {\em Interspeech}, Shanghai, China, Oct. 2020, pp. 5036--5040.

\bibitem{graves2006ctc}
A.~Graves, S.Fern{\'a}ndez, F.Gomez, and J.Schmidhuber,
\newblock ``{Connectionist Temporal Classification: Labelling Unsegmented
  Sequence Data with Recurrent Neural Networks},''
\newblock in {\em ICML}, Pittsburgh, Pennsylvania, USA, June 2006, pp.
  369--376.

\bibitem{lin2017fl}
T.~Lin, P.~Goyal, R.~B. Girshick, K.~He, and P.~Doll{\'{a}}r,
\newblock ``{Focal Loss for Dense Object Detection},''
\newblock in {\em {ICCV}}, Venice,Italy, Oct. 2017, pp. 2999--3007.

\bibitem{zhou2023enhancing}
W.~Zhou, H.~Wu, J.~Xu, M.~Zeineldeen, C.~L{\"u}scher, R.~Schl{\"u}ter, and
  H.~Ney,
\newblock ``{Enhancing and Adversarial: Improve ASR with Speaker Labels},''
\newblock in {\em {ICASSP}}, Rhodes Island, Greece, June 2023, pp. 1--5.

\bibitem{zhang2021orthogonal}
A.~Zhang, A.~Chan, Y.~Tay, J.~Fu, S.~Wang, S.~Zhang, H.~Shao, S.~Yao, and
  R.~Ka{-}Wei Lee,
\newblock ``On orthogonality constraints for transformers,''
\newblock in {\em {ACL/IJCNLP}}, Virtual, Aug. 2021, pp. 375--382.

\bibitem{wang2020orthogonal}
J.~Wang, Y.~Chen, R.~Chakraborty, and S.~X. Yu,
\newblock ``Orthogonal convolutional neural networks,''
\newblock in {\em {CVPR}}, Seattle, WA, USA, June 2020, pp. 11502--11512.

\bibitem{fan2020structuredpruning}
A.~Fan, E.Grave, and A.~Joulin,
\newblock ``{Reducing Transformer Depth on Demand with Structured Dropout},''
\newblock in {\em {ICLR}}, Addis Ababa, Ethiopia, Apr. 2020.

\bibitem{rousseau2014tedlium}
A.~Rousseau, P.~Del{\'e}glise, and Y.~Est{\`e}ve,
\newblock ``{Enhancing the {TED}-{LIUM} Corpus with Selected Data for Language
  Modeling and More {TED} Talks},''
\newblock in {\em LREC}, Reykjavik, Iceland, May 2014, pp. 3935--3939.

\bibitem{vassil2015lbs960}
V.~Panayotov, G.~Chen, D.~Povey, and S.~Khudanpur,
\newblock ``{Librispeech: An {ASR} Corpus based on Public Domain Audio
  Books},''
\newblock in {\em {ICASSP}}, South Brisbane, Queensland, Australia, Apr. 2015,
  pp. 5206--5210.

\bibitem{wei2021phoneme}
W.~Zhou, S.~Berger, R.~Schl{\"{u}}ter, and H.~Ney,
\newblock ``{Phoneme Based Neural Transducer for Large Vocabulary Speech
  Recognition},''
\newblock in {\em {ICASSP}}, Toronto, Canada, June 2021, pp. 5644--5648.

\bibitem{zeyer2018returnn}
A.~Zeyer, T.~Alkhouli, and H.~Ney,
\newblock ``{{RETURNN} as a Generic Flexible Neural Toolkit with Application to
  Translation and Speech Recognition},''
\newblock in {\em {ACL}}, Melbourne, Australia, July 2018, pp. 128--133.

\bibitem{wiesler2014rasr}
S.~Wiesler, A.~Richard, P.~Golik, R.~Schl{\"{u}}ter, and H.~Ney,
\newblock ``{RASR/NN:} the {RWTH} neural network toolkit for speech
  recognition,''
\newblock in {\em {ICASSP}}, Florence, Italy, May 2014, pp. 3281--3285.

\bibitem{lee2021intermediatectc}
J.~Lee, and S.~Watanabe,
\newblock ``{Intermediate Loss Regularization for CTC-Based Speech Recognition},''
\newblock in {\em {ICASSP}}, Toronto, Canada, Jun. 2021, pp. 6224–6228.

\bibitem{louizos2017l0nrom}
C.~Louizos, M.~Welling, and D.~P. Kingma,
\newblock ``{Learning Sparse Neural Networks through $L_0$ Regularization},''
\newblock in {\em {ICLR}}, Canada, Apr. 2018.

\bibitem{xia2022l0norm}
M.~Xia, Z.~Zhong, and D.~Chen,
\newblock ``{Structured Pruning Learns Compact and Accurate Models},''
\newblock in {\em {ACL}}, Ireland, May 2022, pp. 1513--1528.

\bibitem{voita2019att}
E.~Voita, D.~Talbot, F.~Moiseev, R.~Sennrich, and I.~Titov,
\newblock ``Analyzing multi-head self-attention: Specialized heads do the heavy
  lifting, the rest can be pruned,''
\newblock in {\em {ACL}}, Aug. 2019, pp. 5797--5808.

\bibitem{zhang2021att}
S.~Zhang, E.~Loweimi, P.~Bell, and S.~Renals,
\newblock ``{On the Usefulness of Self-Attention for Automatic Speech
  Recognition with Transformers},''
\newblock in {\em {SLT}}, Shenzhen,China, Jan. 2021, pp. 89--96.

\end{thebibliography}
